\documentclass{article}

\usepackage{PRIMEarxiv}

\usepackage[utf8]{inputenc} 
\usepackage[T1]{fontenc}    
\usepackage{hyperref}       
\usepackage{url}            
\usepackage{booktabs}       
\usepackage{nicefrac}       
\usepackage{microtype}      
\usepackage{graphicx}
\usepackage{doi}

\usepackage{cite}
\usepackage{amsmath,amssymb,amsfonts}
\usepackage{algorithmic}
\usepackage{textcomp}
\usepackage{ulem}

\title{
Adversarial Attacks and Defenses in Fault Detection and Diagnosis:\\
A Comprehensive Benchmark on the Tennessee Eastman Process
}

\author{
Vitaliy Pozdnyakov, Aleksandr Kovalenko, Ilya Makarov
\thanks{Corresponding author: Vitaliy Pozdnyakov (e-mail: pozdnyakov@airi.net).
\\
This is accepted to the IEEE Open Journal of the Industrial Electronics Society. DOI: 10.1109/OJIES.2024.3401396.}
\\
AIRI, Moscow, Russia \\
ISP RAS Research Center for Trusted Artificial Intelligence, Moscow, Russia \\
\And
Mikhail Drobyshevskiy, Kirill Lukyanov \\
Ivannikov Institute for System Programming of the Russian Academy of Sciences, Moscow, Russia \\
Moscow Institute of Physics and Technology (National Research University), Moscow, Russia \\
ISP RAS Research Center for Trusted Artificial Intelligence, Moscow, Russia \\
}

\begin{document}
\maketitle

\begin{abstract}

Integrating machine learning into Automated Control Systems (ACS) enhances decision-making in industrial process management. One of the limitations to the widespread adoption of these technologies in industry is the vulnerability of neural networks to adversarial attacks. This study explores the threats in deploying deep learning models for Fault Detection and Diagnosis (FDD) in ACS using the Tennessee Eastman Process dataset. By evaluating three neural networks with different architectures, we subject them to six types of adversarial attacks and explore five different defense methods. Our results highlight the strong vulnerability of models to adversarial samples and the varying effectiveness of defense strategies. 
We also propose a new defense strategy based on combining adversarial training and data quantization. This research contributes several insights into securing machine learning within ACS, ensuring robust FDD in industrial processes.
\end{abstract}


\maketitle

\section{Introduction}
\label{section_introduction}

Automated Control Systems operate with a variety of digital and analog signals received from sensors and control mechanisms. An example is a chemical plant where a set of sensors reflect the condition of an industrial process. 
A common task is Fault Detection and Diagnosis (FDD), where one needs to predict and/or classify a failure based on sensors data. Such methods play a pivotal role in monitoring diverse industrial processes, ranging from chemical processes to electromechanical drive systems. The classification proposed by~\cite{park2020review} categorizes these methods into three groups: those based on expert knowledge, mathematical models, and data-driven approaches. The latter one includes various approaches in machine learning, including neural networks~\cite{sina2023intelligent, winkel2023pseudo, brenner2023better}.
Machine learning algorithms show themselves better than traditional methods based on rules and become more widespread in the area~\cite{BERTOLINI2021114820}. Recent studies demonstrate the success for FDD of various neural network architectures: Multi-Layer Perceptrons, Recurrent Neural Networks, Convolutional Neural Networks~\cite{khoualdia2021optimized,lomov2021fault}.

However, another challenge appears: modern neural networks are vulnerable to adversarial attacks~\cite{liang2022adversarial}. The idea is that the attacker slightly changes the input data (unnoticed) such that the FDD model prediction changes to incorrect. Such attacks are modeled in the literature~\cite{zizzo2020adversarial,kravchik2021efficient}, but it remains unclear if there exist defense strategies good against a wide range of attacks.
In order to address this challenge we benchmark attacks and defenses on the Tennessee Eastman Process dataset~\cite{downs1993plant} where the task is the Fault Detection and Diagnosis (FDD) in a chemical process. 
We consider three various deep learning models --- Multi-Layer Perceptron (MLP), model based on Gated Recurrent Units (GRU), and Temporal Convolutional Network (TCN). We subject these models to six different types of adversarial attacks and explore five defense methods. We analyze the success of these protective measures\footnote{The source code to reproduce our results is available at \url{https://github.com/AIRI-Institute/fdd-defense}.}. Then, a novel protection strategy is proposed, which employs several various defense methods.

The contributions of this work are as follows:
\begin{itemize}
    \item We benchmark popular attack and defense methods on the TEP dataset which shows that existing universal defense methods greatly reduce models quality on original data.
    \item To address this issue we suggest a new defense approach based on adversarial training and data quantization and demonstrate its average effectiveness against various attacks.
    \item We discuss benchmark results and conclude that autoencoders have a potential to be a universal defense methods but they need more research.
\end{itemize}

The rest of the paper is organized as follows. Section \ref{section_review} gives a review on attacks and defense approaches in the area. Section \ref{section_methods} describes methods of attacks and defenses used in our work. In section \ref{section_experiments} we describe experiments and discuss them in section \ref{section_discussion}. Finally, section \ref{section_conclusion} concludes the results.

\section{Review}
\label{section_review}

The operating principles of modern machine learning methods contain vulnerabilities that can be used to carry out various attacks on models. Different methods and areas of application require different adaptations and modifications of an attack developed in one domain area when used in another. Major research into machine learning attack vectors began around 2014 \cite{mozaffari2014systematic}. Over the past five years, this area has advanced very far and many different attack options have been developed.

\subsection{Classification of attacks on machine learning models}

Attacks on machine learning models are typically categorized into several types, which vary based on the capabilities of the attacker in relation to the target and its characteristics throughout the model's lifecycle. These attack types include evasion attacks \cite{biggio2013evasion,teryak2023double,ayub2020model}, poison attacks \cite{jiang2023color, yang2023data, tian2022comprehensive} and exploratory attacks \cite{shokri2017membership, niu2024survey, shaikhelislamov2023study}. Additionally, some of these attack categories have further subgroups. 
Furthermore, attacks are segmented into three groups based on the level of information available about the model's architecture and access to its internal parameters and/or requests: white-box, black-box, and gray-box attacks.


In evasion attacks  (often called adversarial examples) \cite{biggio2013evasion,teryak2023double,ayub2020model}, an attacker interacts with a trained machine learning model and manipulates its behavior by perturbing input samples during testing. The term "evasion" implies that the attacker not only aims to cause the model to behave incorrectly but also seeks to evade detection by both human and automated defense mechanisms. 

Poison attacks \cite{jiang2023color, yang2023data, tian2022comprehensive} are a complex term in the literature. Typically, it refers to injecting poisoned samples into the training dataset with the aim of distorting the training process (so-called data poison attacks). Exploratory attacks involve sending queries to the model to understand its principles of operation. Such attacks can pursue various goals: stealing the model \cite{yuan2024data, zhang2023apmsa}, conducting membership inference attacks \cite{shokri2017membership, niu2024survey, shaikhelislamov2023study}, and others. In this article the main focus was on evasion attacks, as they pose the greatest threat due to not requiring an insider attacker and directly impacting the model's predictions.

In white-box attacks \cite{wang2022di, singhal2023enhancing}, the attacker possesses complete information, enabling them to execute any operations on their instance of the deployed model (e.g., obtaining gradients, accessing output data from any layer) to construct perturbed samples. In the presence of defense mechanisms, these mechanisms are also susceptible to attack by the adversary.

Black-box attacks \cite{bhagoji2017exploring, zhang2020deepsearch, bai2023query} assume that the attacker can only make a limited number of requests ($L$, where $1 \leq L < \infty$) to the deployed model. The models provide the attacker with predictions such as label class or probability, semantic map segmentation, etc. 

Gray box (or semi-white box) \cite{xiang2020side, al2023incremental} attacks represent an intermediate state between white-box and black-box scenarios. They involve the imposition of certain restrictions that provide some information about the learning model/process, albeit incomplete.

\subsection{Most common methods of attack and defense}
In the literature, there are numerous articles that explore issues akin to those addressed in this study. However, these methods are either examined on different datasets or pertain to disparate domain areas, necessitating adaptation for our purposes. Notably, many initial attacks were devised for image analysis models, and not all methods have been fully tailored for the domain area under our investigation. Therefore, in this review, we also consider these aspects.
Articles such as \cite{miller2020adversarial, aldahdooh2022adversarial, laykaviriyakul2023collaborative, hou2022similarity, zizzo2019adversarial} delve into the impact of attacks on various image datasets like CIFAR-10, CIFAR-100, ResNet-20, and MNIST. These articles also discuss various protective measures. Key attack methods include L-BFGS, FGSM, PGD, C\&W, and DeepFool. While identifying the most prevalent defense methods can be challenging, several key approaches emerge, notably Defense-GAN and Adversarial Training.
Moreover, articles such as \cite{zizzo2020adversarial, specht2018generation, kravchik2021efficient, hamdi2020sada, gomez2021crafting, gomez2022methodology, zhuo2022attack} discuss attacks and defense methods on datasets like TEP and/or similar domain areas such as CARLA, Electra, SWaT, BATADAL, and WADI. Many attack and defense methods in these articles share operational principles with those used in computer vision.


In white-box adversarial attacks, access to gradients is a primary tool. Attackers exploit gradients by calculating the gradient of the loss function concerning the input. Then, they perturb the input in the direction of the gradient to maximize the loss. 
The mathematical details of the attack algorithms that will be used in this article are outlined in the next section (section \ref{section_methods}).

The Fast Gradient Sign Method (FGSM), proposed by \cite{Goodfellow2014ExplainingAH}, generates adversarial examples with a single gradient step. It updates the input based on the direction of increasing loss, using a small multiple of the sign of the gradient. While FGSM is fast, its success rate for adversarial examples is low.

To improve the success rate of FGSM, \cite{madry2017towards} introduced the Projected Gradient Descent (PGD) method. Unlike FGSM, PGD takes multiple smaller steps in the gradient direction and clips the result by a specified value. Although PGD is more effective than FGSM in finding adversarial examples closer to the model's decision boundary, it is computationally more expensive due to requiring multiple iterations.

For untargeted attacks, \cite{moosavi2016deepfool} proposed the DeepFool method optimized for the $L_2$ distance metric. It assumes the linearity of the decision boundary in neural networks and finds the minimum adversarial perturbation needed to fool the classifier. DeepFool iteratively identifies the direction that maximally changes the current prediction of the neural network and takes a small step in that direction until finding a true adversarial example.

A more sophisticated white-box attack, the C\&W attack by \cite{Carlini2016TowardsET}, is applicable under various distance metrics: $L_0$, $L_2$, and $L_\infty$. This attack optimizes a loss function considering the distance between the original input and the adversarial example, along with the classifier's prediction confidence. The optimization includes a constraint on the perturbation size, making the resulting adversarial example more realistic and challenging to detect.


Adversarial training \cite{miller2020adversarial, sinha2018gradient, balunovic2019adversarial} is a widely used defense technique aimed at making neural networks more resilient to adversarial attacks. Instead of relying solely on traditional training data, adversarial training incorporates examples with adversarial biases into the training process. Adversarial training has been shown to be effective in improving the robustness of neural networks to various types of adversarial attacks, including both white-box and black-box attacks. Despite some problems, adversarial training remains one of the most effective methods for protecting against adversarial attacks and is widely used in practice to improve the security and reliability of neural networks. 

\subsection{Interactions among defense methods}

While the literature offers many different methods for defending machine learning models, there are few studies that explore building models combining multiple defense methods. In the paper \cite{szyller2023conflicting} the authors examine the possibility of combining the most popular defense methods against evasion and poisoning attacks. The research concludes that many methods, at the level of algorithmic ideas, are incompatible, demonstrating this through practical examples. Therefore, constructing models that combine defense methods is a complex and underexplored task.

\subsection{Summary of the review}

Since the main objective of the article is to create a benchmark,  we pay particular attention  to the most common methods described in the literature. This research focuses on analyzing vulnerabilities and implementing protection strategies within ACS. The following section elaborates on the mathematical aspects of the methods employed in this research.

\section{Methods}
\label{section_methods}
\subsection{Fault diagnosis methods}

Fault detection and diagnosis (FDD) methods, are widely used in monitoring industrial processes, such as chemical processes \cite{reinartz2021extended} and electromechanical drive systems \cite{lessmeier2016condition}. The authors of \cite{park2020review} divide FDD methods into three groups: data-driven, model-based, and knowledge-based approaches. In our work, we investigate the properties of data-driven methods.

Data-driven FDD problem is formulated as follows. Let there be a sequence of observations $X_1, \dots, X_n$, where $X_t \in R^d$ are the values of sensors at time $t$. Thus, $X_1,\dots,X_n$ form a multivariate time series. Also, let there be a sequence of labels $y_1,\dots,y_n$ where $y_t \in \{0,1\}^m$ defines the type of fault at time $t$. If $\arg\max(y_t)=0$, the process is in the normal state, otherwise $\arg\max(y_t)$ determines the fault number. Then for a sliding window of width $k$, we need to find such a function $f:R^{d \times k} \to [0,1]^m$ that $$f=\arg\min_f \frac{1}{n-k} \sum_{t=k}^n l(y_t,f(X_{t-k+1},\dots,X_t )),$$ where $l$ is some loss function, most commonly cross-entropy, also known as Log Loss. The function $f$ can be found using machine learning methods.

In recent years, many deep learning methods based on different neural network architectures were proposed to solve FDD problem. The simplest one is MLP that was applied to FDD in \cite{khoualdia2021optimized, ali2019machine, unal2014fault, ghate2010optimal}. Multivariate time series is converted to a vector of concatenated observation and then processed by MLP to predict the process state. TCN is another popular architecture for FDD \cite{li2021novel, zhang2022real, zhang2021ms}. TCN is a modification of a 1D convolutional network with causal and dilated convolutions \cite{bai2018empirical} that helps to process sequential data with long-term dependencies. In addition, GRU is a type of recurrent neural networks that shows SOTA results of FDD on many datasets including Tennessee Eastman Process \cite{yuan2019intelligent, lomov2021fault}.

\subsection{Adversarial attacks}
During the attack, an adversarial sample $X_t^{'}$ is created such that: $f(X_t^{'}) \neq f(X_t)$, where $X_t^{'} = X_t + \mathcal{N}$ and $\mathcal{N} \in R^{d \times k}$ is a perturbation matrix. Strength of an attacks is defined by the maximal shift $\epsilon$ as follows: $\|X_t-X_t^{'}\|_\infty \leq \epsilon$. 

When choosing types of attacks, we proceeded from the assumption that the attacker has access to either only input and output data or all information about the data and model architecture. 2 black-box (Random noise, FGSM distillation) and 4 white-box attacks (FGSM, PGD, DeepFool, Carlini and Wagner) were implemented.

\subsubsection{Random Noise}
Random noise is the simplest black-box attack based on adding random values to the input data:
\begin{align*}
x^{'}=x+\epsilon z
\end{align*}
where $\epsilon$ limits the magnitude of noise values and $z$ is distributed according to Bernoulli’s principle with parameter $p=0.5$ on the sample space of elementary events $\{-1,1\}$.

\subsubsection{Fast Gradient Sign Method (FGSM)}
FGSM \cite{Goodfellow2014ExplainingAH} is a white-box attack based on the gradient of the loss function calculated for the input data. The signs of obtained gradient vector indicate the direction in which the input data should be changed to increase the probability of model error. The attack consists of shifting each value of the data by a step of size $\epsilon$, with a sign corresponding to the gradient:
\begin{align*}
x^{'}=x+\epsilon \operatorname{sign}[\triangledown l(f(x),y)]
\end{align*}

\subsubsection{FGSM Distillation}
Distillation can be used to create a black-box adversarial attack as proposed in \cite{cui2020substitute}. Based on the input and output data of the model, a neural network classifier with an arbitrary architecture can be trained. Adversarial samples are obtained by attacking the resulting model by any white-box attack. In our study we used MLP architecture and FGSM attack.

\subsubsection{Projected Gradient Descent (PGD)}
PGD \cite{madry2017towards} is an iterative modification of the FGSM white-box attack method. The main difference is that the data shift is done in several steps. After each step, the gradient signs are recalculated:
\begin{align*}
x_{i+1}^{'}=\operatorname{Clip}_\epsilon \{x_i^{'}+\alpha \operatorname{sign[\triangledown l(f(x),y)]}\}
\end{align*}
where $x_i^{'}$ denotes the changed input data since the previous iteration, $\operatorname{Clip}\{\}$ limits the resulting data shift to no more than $\epsilon$, and $\alpha$ denotes the shift step size at each iteration.

\subsubsection{DeepFool}
DeepFool \cite{moosavi2016deepfool} is a white-box attack which minimizes the difference between the elements of the output vector $f(x)$ that correspond to the correct and incorrect fault type. Among all possible incorrect types, the closest in absolute value of the difference is selected. Minimization occurs in several steps, each defined as follows:
\begin{align*}
x_{i+1}^{'}=x_i+\frac{|D(x_i)|}{||\triangledown D(x_i)||_1} \triangledown D(x_i)
\end{align*}
where $D(x)=f(x)_\text{false}-f(x)_\text{true}$. $f(x)_\text{false}$ is the value of the output vector corresponding to the nearest incorrect fault type, which is selected independently at each step. After each iteration, the total adversarial vector $x_{i+1}^{'}$ is limited by $\epsilon$ value.

\subsubsection{Carlini and Wagner (C\&W)}
C\&W \cite{Carlini2016TowardsET} is a white-box attack that minimizes the sum of the shift value over the distance metric $D$ and the value of some auxiliary function $g$. Function $g$ takes negative values in case of incorrect classification. This optimization problem can be represented as:
\begin{align*}
\text{min}_\eta D(x, x+\eta)+g(x+\eta)
\end{align*}
where $g(x)=\text{ReLU}(\arg \max(y)-\arg \max(f(x)))$ and $D$ is Chebyshev distance. Minimization is performed by the stochastic gradient descent method or its analogues. For comparison with other attack methods, we constrain $\eta$ according to the selected $\epsilon$ value.

\subsection{Defense methods}
Another goal of the study was to find out how defense methods behave under attacks with different strengths and for different neural network architectures. The five most popular strategies were implemented: Adversarial training, Autoencoder, Quantization, Regularization, and Distillation. We also proposed to protect models by combination of defense methods.

\subsubsection{Adversarial Training}
Adversarial training method \cite{Goodfellow2014ExplainingAH} consists of adding adversarial samples to the training set. The training loss function is given as follows:
\begin{align*}
L=l(f(x),y)+\lambda l(f(x^{'}),y)
\end{align*}
where $x^{'}$ is adversarial sample and $\lambda$ is adversarial training coefficient.

\subsubsection{Defensive Autoencoder}
Autoencoder can be used to reconstruct attacked data as proposed in \cite{meng2017magnet}. During its training, the following loss function is minimized:
\begin{align*}
L=||x_{AE}-x||_1
\end{align*}
where $x_{AE}=\text{autoencoder} (x + \varepsilon)$ is a reconstructed data and $\varepsilon$ is added noise. 

\subsubsection{Data Quantization}
Quantization is a preprocessing method that converts continuous values into a set of discrete values on a uniform grid \cite{guo2017countering}. This approach reduces the quality of the input data but can neutralize the impact of adversarial attacks. The fault diagnosis model must be retrained on quantized data.

\subsubsection{Gradient Regularization}
The fault diagnosis model can be protected by training using gradient regularization \cite{finlay2019scaleable} of the loss function over the input data:
\begin{align*}
L=l(f(x),y)+\lambda \left( \frac{1}{h^2 n} ||f(z)-f(x)||_2^2 \right)
\end{align*}
where
\begin{align*}
z=x+h \frac{\triangledown l(f(x),y)}{||\triangledown l(f(x),y)||_2},
\end{align*}
$h$ is a  quantization step and $\lambda$ is a regularization coefficient.

\subsubsection{Defensive Distillation}
Distillation defense method \cite{papernot2016distillation} refers to the process of creating a copy of the original neural network model that is more resistant to adversarial attacks. The original neural network is called the teacher, and the new neural network is called the student. When teaching a student, so-called smooth labels are used, which are obtained using the activation function $\operatorname{softmax}(x, T)$ on the last layer of the teacher:
\begin{align*}
\operatorname{softmax}(x, T)_i=\frac{e^{x_i/T}}{\sum_je^{x_j/T}},
\end{align*}
where $T$ is a temperature constant. At $T=0$ the function converges to a maximum and at $T\rightarrow\infty$ the function converges to a uniform distribution.

\subsubsection{Adversarial Training on Quantized Data}
In recent years, a lot of research has been carried out to develop new defense methods against adversarial attacks. New ideas emerge that are superior to previous approaches in certain conditions. However, there is still no ideal defense method capable of protecting against all types of threats. The vulnerability of protected neural networks is reduced only under certain types of adversarial attacks; in other cases, the accuracy of the models drops noticeably.


In this paper, we propose to use a combination of adversarial training and data quantization. As was shown in \cite{guo2017countering}, quantization allows to clean the input from adversarial perturbation due to the grid alignment of discrete values. However, the size of the grid (quantization frequency), has an important role in this type of protection. If the grid is too wide, it reduces the quality of fault diagnosis, if the grid is too narrow, only a fraction of the data can be effectively recovered. On the other hand, adversarial training provides high model robustness but reduces the quality of diagnosis. This happens because during training, the data contains many adversarial examples that degrade the model's ability to generalize important dependencies in the data that help diagnose faults. Thus, at high values of $\epsilon$ in adversarial training, the quality of the model drops significantly, otherwise it does not provide a sufficient level of protection. 

We propose to use adversarial training on the data after quantization. Thus, during training, we attack the data with an adversarial attack such as FGSM. We then quantize this data and feed it into the input of the model as a training set. Quantization allows to reduce the strength of the attack, which in turn allows the model to generalize better during adversarial training. As a result, quantization helps the model to achieve better quality in adversarial training.

An additional advantage of this approach is that it does not require a separate model, as is the case with the distillation method or the autoencoder. It is also quite efficient in terms of computational time and memory, since quantization takes place in linear time and requires no additional memory, while adversarial training has the same complexity as training a model on the original data and also requires no additional memory.

\subsection{Dataset}

The Tennessee Eastman Process is a very popular dataset for benchmarking fault detection and diagnosis methods. It describes the operation of a chemical production line, where the process smoothly transitions from a normal state to a faulty one. In our study, we used a version of the TEP extended by Reinartz et al.~\cite{reinartz2021extended} that contains significantly more sensor data than the original (5.2 Gb vs 58 Mb). This version includes 100 simulation runs for each of the 28 fault types. Each run consists of 52 sensor values for 2000 timestamps, and thus the input samples are in the form of matrices $X^{k \times 52}$, where $k$ is the sliding window size. All data in our experiments were standardized by removing the mean and scaling to unit variance.

\section{Experiments}
\label{section_experiments}

In our study, we wanted to find out how adversarial attacks affect FDD models based on neural networks with different architectures and what defense methods can be used. To analyze the impact of adversarial attacks on fault diagnosis models, the accuracy metric was chosen. This metric well reflects changes in the quality of models when the data is attacked.

The description of our experiments is divided into 4 subsections. The FDD models subsection describes the training process of neural networks with different architectures. The next subsection shows how the accuracy of the models changes under different types of attack. Further, various methods for protecting models and their properties are shown. Finally, on the basis of the results of experiments, we also proposed and evaluated an approach consisting of a combination of two defense methods. All final results can be found in Fig. \ref{fig:all} and Tables \ref{table:mlp}-\ref{table:tcn}.

\subsection{FDD models}

For our experiments, we used three models of neural networks with different architectures. To make the models differ from each other more, they contain different numbers of parameters and were trained for different numbers of epochs. The first model is a multilayer perceptron (MLP) consisting of two linear layers and containing 3 452 949 parameters. The second one is based on recurrent gated units (GRU) and containing 204 565 parameters. We also used temporal convolutional network (TCN) with 151 935 parameters. Data were standardized with a standard deviation of 1. Sliding window size was 32, which is a compromise between the accuracy of the models and the duration of the experiments. All models were trained on the TEP dataset for 20, 5 and 10 epochs for MLP, GRU and TCN respectively. The accuracy metrics for fault diagnosis on non-attacked data are presented in Table \ref{table:1}. Combinations of the number of parameters and training epochs are selected on the validation set.

\begin{table}[!ht]
\caption{Accuracy of unprotected models on normal data.}
\label{table:1}
\centering
\begin{tabular}{ c|c } 
  Model & Accuracy \\ 
 \hline
 MLP & $0.8873 \pm 0.0002$ \\
 GRU & $0.9067 \pm 0.0041$ \\
 TCN & $0.8985 \pm 0.0097$ \\
\end{tabular}
\end{table}

The selected neural network architectures showed similar accuracy and can effectively solve the fault diagnosis task.

\subsection{Attacks on FDD models}

\begin{figure}[!ht]
    \centering
    \includegraphics[width=0.45\textwidth]{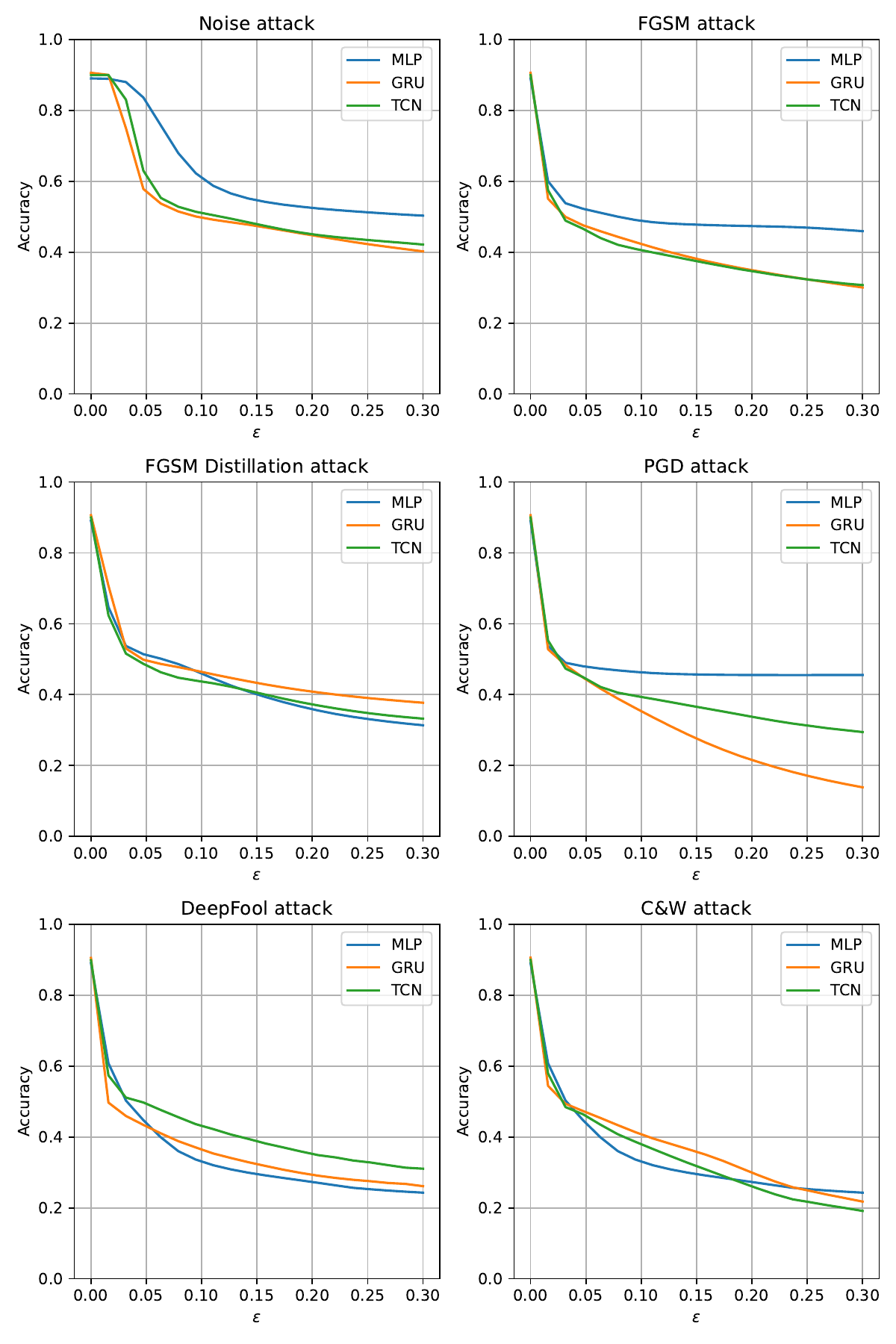}
    \caption{Accuracy drop of unprotected models under six different types of attacks depending on the strength of an attack $\epsilon$.}
    \label{fig:Unprotected}
\end{figure}

At the next stage, unprotected models were attacked by six types of attacks with different $\epsilon$. For $\epsilon$ values, 20 points were selected in the range from 0 to 0.3 with a step of 0.015. We consider this range to be reasonable given that the data is scaled to a unit variance and the attack should not be detected  by both human and automated defense mechanisms. Fig. \ref{fig:Unprotected} shows how the model's accuracy degrades depending on the type and strength of the attack. It decreases significantly with small shifts in the attacked data for $\epsilon$ values less than 0.05.

To cause potential harm, an attacker does not always need to have access to model architectures and use white-box attacks. Experiments have shown that to create a strong adversarial attack, it is enough to have access to the input and output data of the FDD system. This data can be used to train an arbitrary neural network architecture on the basis of which adversarial samples will be created. The distillation FGSM attack showed a similar effect on the accuracy of models as white-box attacks in our study. This type of black-box attacks seems to be the easiest to carry out and potentially the most dangerous.

\subsection{Protection of FDD models}

All three neural network architectures have proven to be highly vulnerable to adversarial attacks and require protection methods. The defense methods studied in our research have many variations and parameters for selecting. It is not possible to conduct experiments for all combinations of settings and models in an adequate period of time. Therefore, we took only the TCN model, which has the fastest inference, to select more optimal settings for defense methods adjustment. Experiments conducted for each type of protection are described in following subsections. After setting up, the defense methods were applied to all FDD models and the final results are presented in Fig. \ref{fig:all} and Tables \ref{table:mlp}-\ref{table:tcn}.

\subsubsection{Adversarial Training}

\begin{figure}[!ht]
    \centering
    \includegraphics[width=0.45\textwidth]{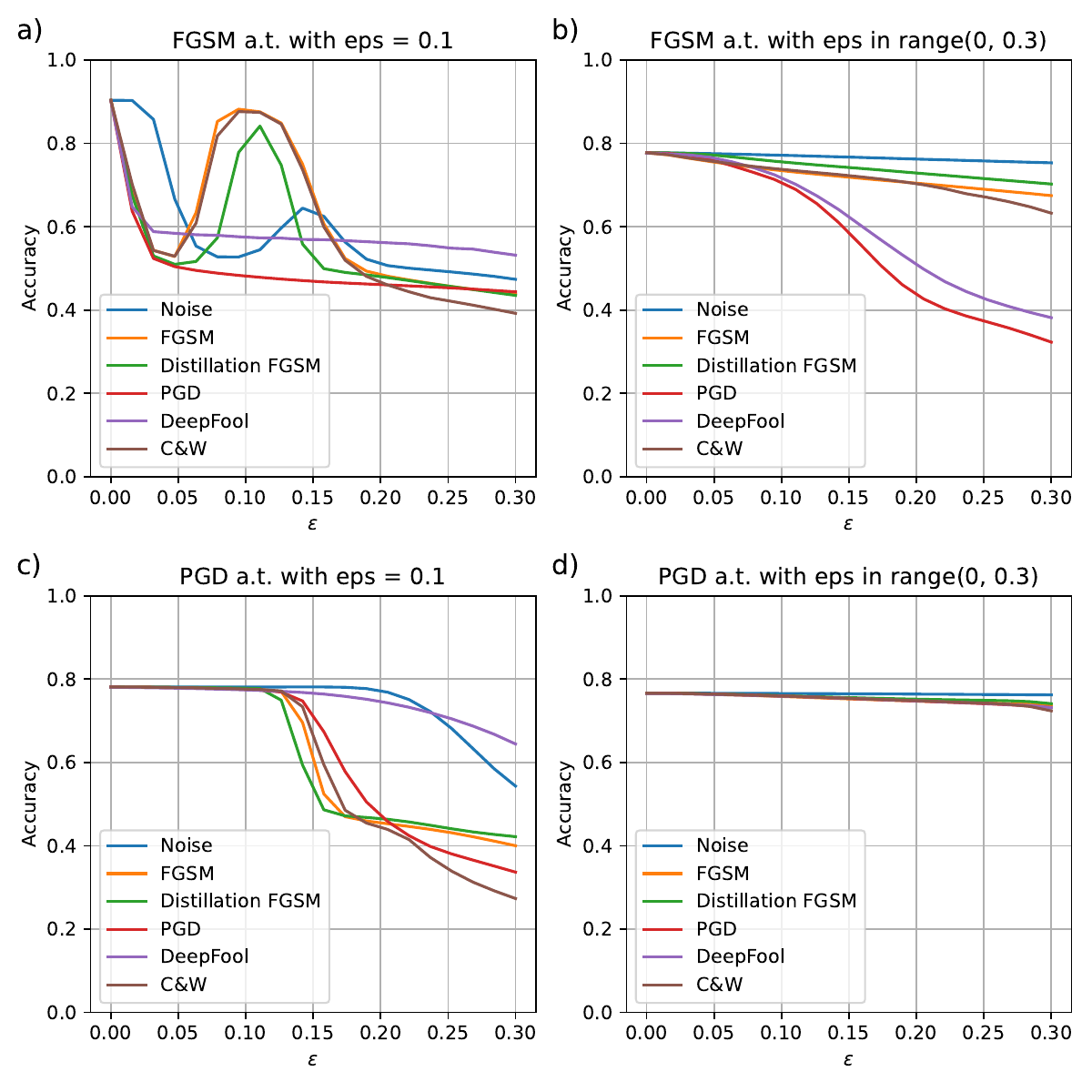}
    \caption{Accuracy of the TCN model protected by adversarial training with different settings: a) training on FGSM samples with fixed $\epsilon = 0.1$; b) training on FGSM samples with set of $\epsilon$ values from the range (0, 0.3); c) training on PGD samples with fixed $\epsilon = 0.1$; d) training on PGD samples with set of $\epsilon$ values from the range (0, 0.3).}
    \label{fig:adv_tr}
\end{figure}

In our study we used equal amounts of normal and attacked data for adversarial training method. Experiments have shown a strong dependence of the model robustness on the set of adversarial samples during the training process. As an example the model trained on attacked data with $\epsilon$ value 0.1 is not protected from attacks with $\epsilon$ values 0.05 and 0.2. Fig. \ref{fig:adv_tr} shows changes in the TCN model's accuracy after adversarial training with different options. The first one is the training with FGSM adversarial samples and fixed $\epsilon$ value equal to 0.1. Then the number of $\epsilon$ values was expanded to the set with range from 0.015 to 0.3 ($\epsilon$ values were randomly selected from the set for every data sample). The same measurements were made for training with PGD adversarial samples.

Adding more different perturbed data to the training process increases the average robustness of the model to adversarial attacks. But the quality with normal data decreases. Table \ref{table:at} shows the accuracy of the TCN model on non-attacked data before and after adversarial training. Training with PGD adversarial samples showed better robustness from all attack types but worse quality in non-attacked mode. We used this setting for the final comparison of all defense methods.

\begin{table}[!ht]
\caption{Accuracy of the TCN model protected by adversarial training on normal data.}
\label{table:at}
\centering
\begin{tabular}{ c|c } 
 Type of adversarial training & Accuracy \\ 
 \hline
 None & 0.90 \\
 FGSM with $\epsilon$ = 0.1 & 0.90 \\
 FGSM with $\epsilon$ in range(0, 0.3) & 0.78 \\
 PGD with $\epsilon$ = 0.1 & 0.78 \\
 PGD with $\epsilon$ in range(0, 0.3) & 0.77 \\
\end{tabular}
\end{table}

\subsubsection{Defensive Autoencoders}

\begin{figure}[!ht]
    \centering
    \includegraphics[width=0.45\textwidth]{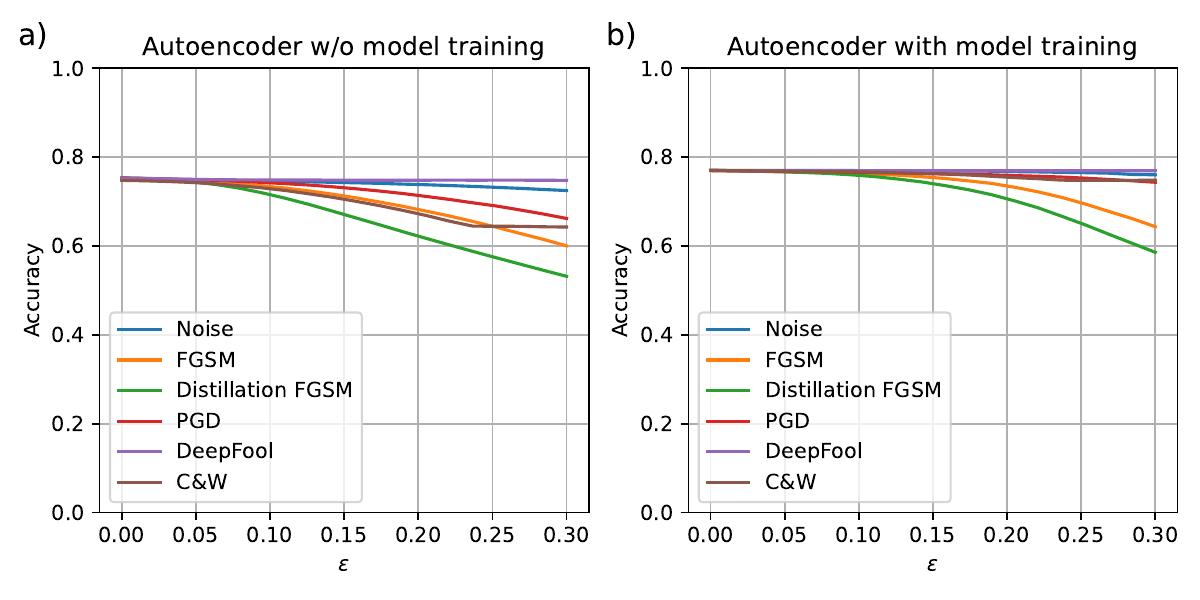}
    \caption{Accuracy of the TCN model protected by autoencoder: a) model was trained on the original data; b) model was trained on the data obtained at the output of autoencoder.}
    \label{fig:ae}
\end{figure}

During the experiments, we trained a simple autoencoder with linear layers in the encoder and decoder parts. There are two options for using it in conjunction with the models. The model can be trained on the original dataset data or on autoencoder output data. Both approaches are shown in Fig. \ref{fig:ae} using the TCN model as an example. Experiments have shown that when using an autoencoder, a model trained on its output shows better quality and robustness to adversarial attacks. This setting was chosen for the final comparison for all models.

Accuracy metrics on non-attacked data can be seen in Table \ref{table:ae}. It is significantly lower than on unprotected model, but there is an opportunity to experiment with advanced autoencoder architectures in further research.

\begin{table}[!ht]
\caption{Accuracy of the TCN model protected by autoencoder on normal data.}
\label{table:ae}
\centering
\begin{tabular}{ c|c } 
 Type of autoencoder defense & Accuracy \\ 
 \hline
 None & 0.90 \\
 Autoencoder w/o model training & 0.75 \\
 Autoencoder with model training & 0.77 \\
\end{tabular}
\end{table}

\subsubsection{Data Quantization}

\begin{figure}[!ht]
    \centering
    \includegraphics[width=0.45\textwidth]{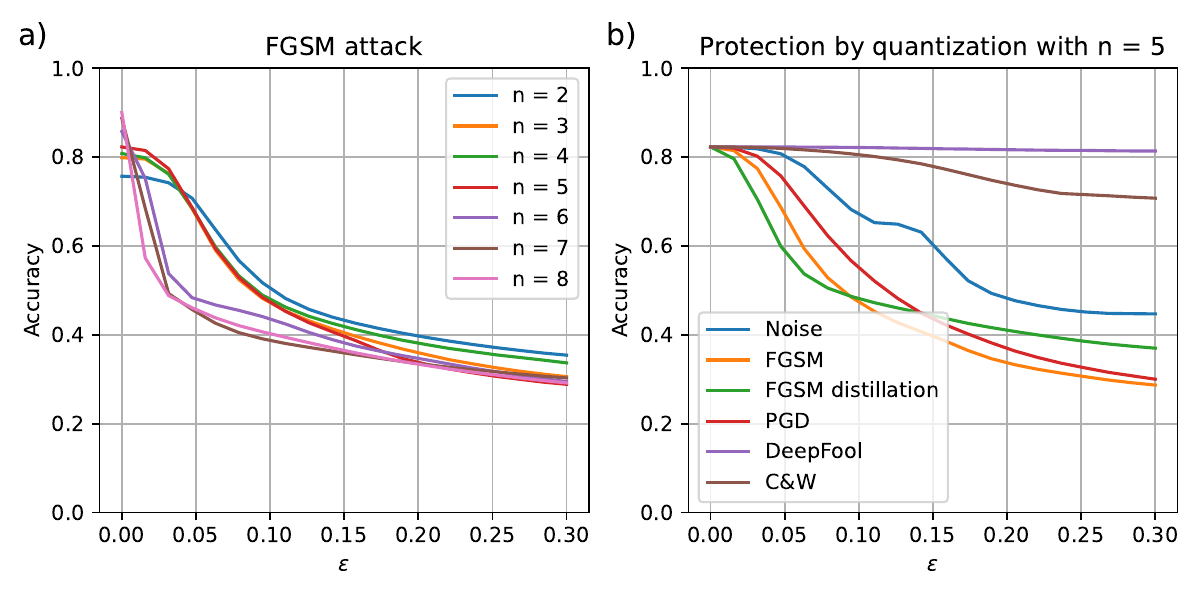}
    \caption{Accuracy of the TCN model protected by quantization: a) model is under FGSM attack and $n$ indicates the number of discrete values during the quantization process ($2^n$); b) model is protected by quantization with $n=5$ under six types of attacks.}
    \label{fig:quant}
\end{figure}

Quantization converts continuous input data into a set of discrete values. To select the number of discrete values in the set, we used different values $n$ for powers of two (from $2^2$ to $2^8$). The left part of Fig. \ref{fig:quant} shows the accuracy of the TCN model protected by quantization method with different sets of discrete values. The attacks where made by FGSM adversarial samples. The remaining types of attacks on the model protected by quantization with $n=5$ are presented on the right side of the picture.

Table \ref{table:quant} shows the accuracy of the TCN model protected by quantization with different numbers of discrete values on non-attacked data.

\begin{table}[!ht]
\caption{Accuracy of the TCN model protected by quantization with different parameter $n$ on normal data.}
\label{table:quant}
\centering
\begin{tabular}{ c|c } 
 Number of discrete values $= 2^n$ & Accuracy \\ 
 \hline
 None & 0.90 \\
 n = 2 & 0.76 \\
 n = 3 & 0.80 \\
 n = 4 & 0.81 \\
 n = 5 & 0.82 \\
 n = 6 & 0.86 \\
 n = 7 & 0.87 \\
 n = 8 & 0.90 \\
\end{tabular}
\end{table}

For the final comparison, the setting with $n=5$ was chosen as a trade-off between the model’s robustness to attacks and the accuracy on normal data.

\subsubsection{Gradient Regularization}

\begin{figure}[!ht]
    \centering
    \includegraphics[width=0.45\textwidth]{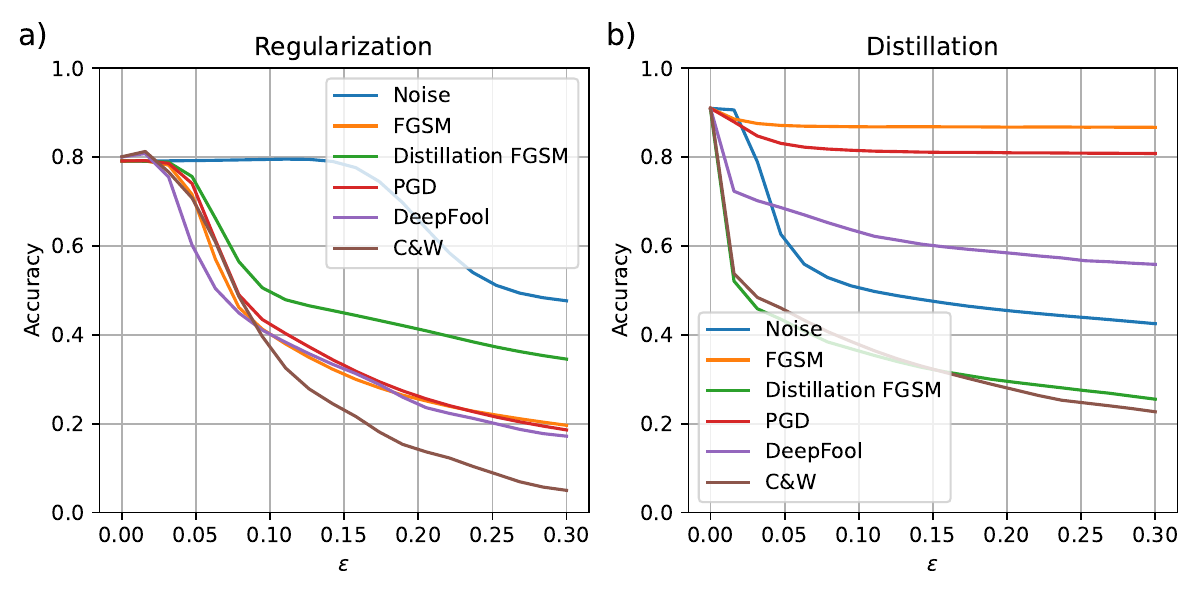}
    \caption{Accuracy of the TCN model protected by: a) regularization defense method; b) distillation defense method.}
    \label{fig:reg}
\end{figure}

The parameters for tuning the regularization method did not show a significant impact on the effectiveness of the protection. The Fig. \ref{fig:reg} (a) shows the change in the accuracy of the protected TCN model after all types of attacks. The quantization step parameter $h$ and regularization coefficient $\lambda$ were equal to 0.001 and 1 respectively. Regularization turned out to be useful just for small $\epsilon$ values. However, it well improves robustness against random noise.

\subsubsection{Defensive Distillation}

Distillation is a gradient masking technique that protects models against gradient-based adversarial attacks. Changing the temperature constant parameter $T$ does not significantly affect the effectiveness against other types of attacks. The Fig. \ref{fig:reg} (b) shows the accuracy of the TCN model protected by distillation defense method with parameter $T=100$. The results confirm good protection against gradient-based FGSM and PGD adversarial attacks. However, other threats remain relevant when using this protection method.

\subsubsection{Adversarial Training on Quantized Data}

The experimental results showed that various defense methods can be effective against some types of attack and not against the others. This fact suggests the idea of using several defense approaches together. In our study, we used a combination of adversarial training and quantization defense methods. 

\begin{figure}[!ht]
    \centering
    \includegraphics[width=0.45\textwidth]{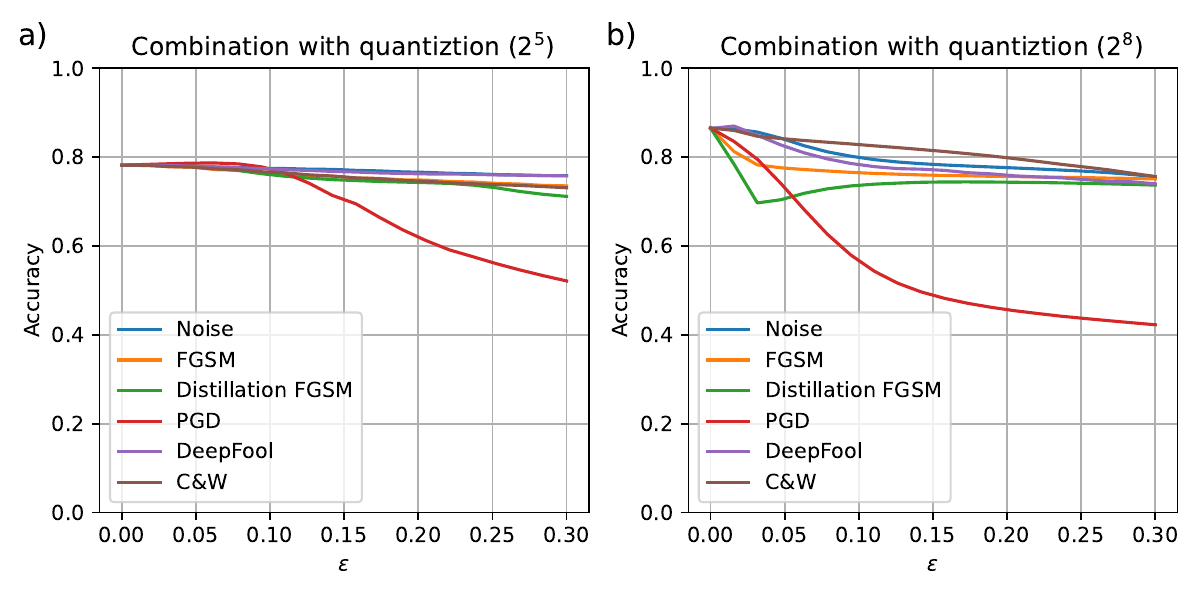}
    \caption{Accuracy of the TCN model protected by combination of FGSM adversarial training ($\epsilon=0.1$) and: a) quantization having $2^5$ discrete values; b) quantization having $2^8$ discrete values.}
    \label{fig:comb}
\end{figure}

For the adversarial training setting we chose attack with FGSM samples and $\epsilon=0.1$ (Fig. \ref{fig:adv_tr} (a)). We combined it with the quantization having $2^5$ discrete values (Fig. \ref{fig:quant} (a)). The results of this combination significantly exceed the effectiveness of these methods separately (Fig. \ref{fig:comb} (a)). Moreover, we tried to change the quantization defense setting by increasing the number of discrete values to $2^8$. Its combination with FGSM adversarial traning are shown in Fig. \ref{fig:comb} (b).

Quantization defense method having $2^8$ discrete values is more vulnerable to adversarial attacks than the one having $2^5$ values. But its combination with adversarial training gives partially better results, especially on normal, non-attacked data (Table \ref{table:comb}). 

\begin{table}[!ht]
\caption{Accuracy of the TCN model protected by combination of adversarial training and quantization defense method on normal data.}
\label{table:comb}
\centering
\begin{tabular}{ c|c } 
 Type of combination & Accuracy \\ 
 \hline
 None & 0.90 \\
 Adv. tr. and quantization with $n=5$ & 0.79 \\
 Adv. tr. and quantization with $n=8$ & 0.89 \\
\end{tabular}
\end{table}

\section{Discussion}
\label{section_discussion}

Our experiments confirmed the vulnerability of fault diagnosis models based on different neural networks to adversarial attacks. We implemented six types of attacks and all of them lead to a significant decrease in accuracy of FDD methods. However, a good defense method should be effective against any type of adversarial attack. At the same time, the accuracy of defended models should not drop significantly on normal, non-attacked data. The $\epsilon$ parameter, which limits the maximum shift in the attacked data, is common to all types of attacks. The choice of $\epsilon$ value range when creating protection for models depends on many factors (such as additional systems for detecting adversarial attacks) and is the subject of discussion. In our work, we investigated five types of defense methods against adversarial attacks with $\epsilon$ values in the range (0, 0.3). We also proposed a combination of adversarial training and quantization defense methods.

Adversarial training with PGD samples and defense by autoencoder can be considered as universal methods against adversarial attacks over a wide range of $\epsilon$ values. The disadvantage of these approaches is a significant decrease in accuracy on normal non-attacked data. Adversarial training can be done against the attack with a specific $\epsilon$ value without losing accuracy on normal data but will be ineffective for attacks with other $\epsilon$ values. Adding more variety to adversarial samples degrades the overall accuracy of the model.

The accuracy of the model protected by autoencoder on non-attacked data has noticeably decreased, but was stable after most types of attacks. This approach seems to have great potential and requires further research with different autoencoder architectures. Additionally, the vulnerability of the autoencoders themselves should be studied. The disadvantage of this method is the need for additional computing resources.

Other methods such as quantization, regularization and distillation have shown high protection against some types of attacks and poor results against the others. To address the limitations of individual defense methods, we explored the possibility of combining them. In our study, we combined FGSM adversarial training and quantization defense method. This approach provides good protection against most types of attacks (except for PGD adversarial examples with large $\epsilon$ values) with small losses in quality. It is computationally efficient and does not require additional memory. Other combinations of various defense methods can be explored in further research.


\section{Conclusion}
\label{section_conclusion}
This study confirmed that adversarial attacks can greatly reduce the quality of FDD models. Such attacks can be quite feasible if attackers have access to the data exchange system. Therefore it is important to know the robustness of models used in real systems to adversarial samples. There are many types of attacks and quite a few universal defense methods capable of protecting against all of them simultaneously. Also universal defense methods significantly reduce the accuracy of the models on normal non-attacked data. Here it seems that autoencoders can be improved in further research through the use of more advanced architectures. Many of defense methods protect against certain types of adversarial attacks with a good efficiency. Such methods can be combined into one more powerful defense system, like our experiment with adversarial training and quantization.


\section*{Acknowledgment}

This work was supported by a grant for research centers in the field of artificial intelligence, provided by the Analytical Center 
in accordance with the subsidy agreement (agreement identifier 000000D730321P5Q0002) and the agreement with the Ivannikov Institute for System Programming of the Russian Academy of Sciences dated November 2, 2021 No. 70-2021-00142.

\section*{APPENDIX A. Comparison of all methods}

\begin{figure*}[!ht]
    \centering
    \includegraphics[width=0.64\textwidth]{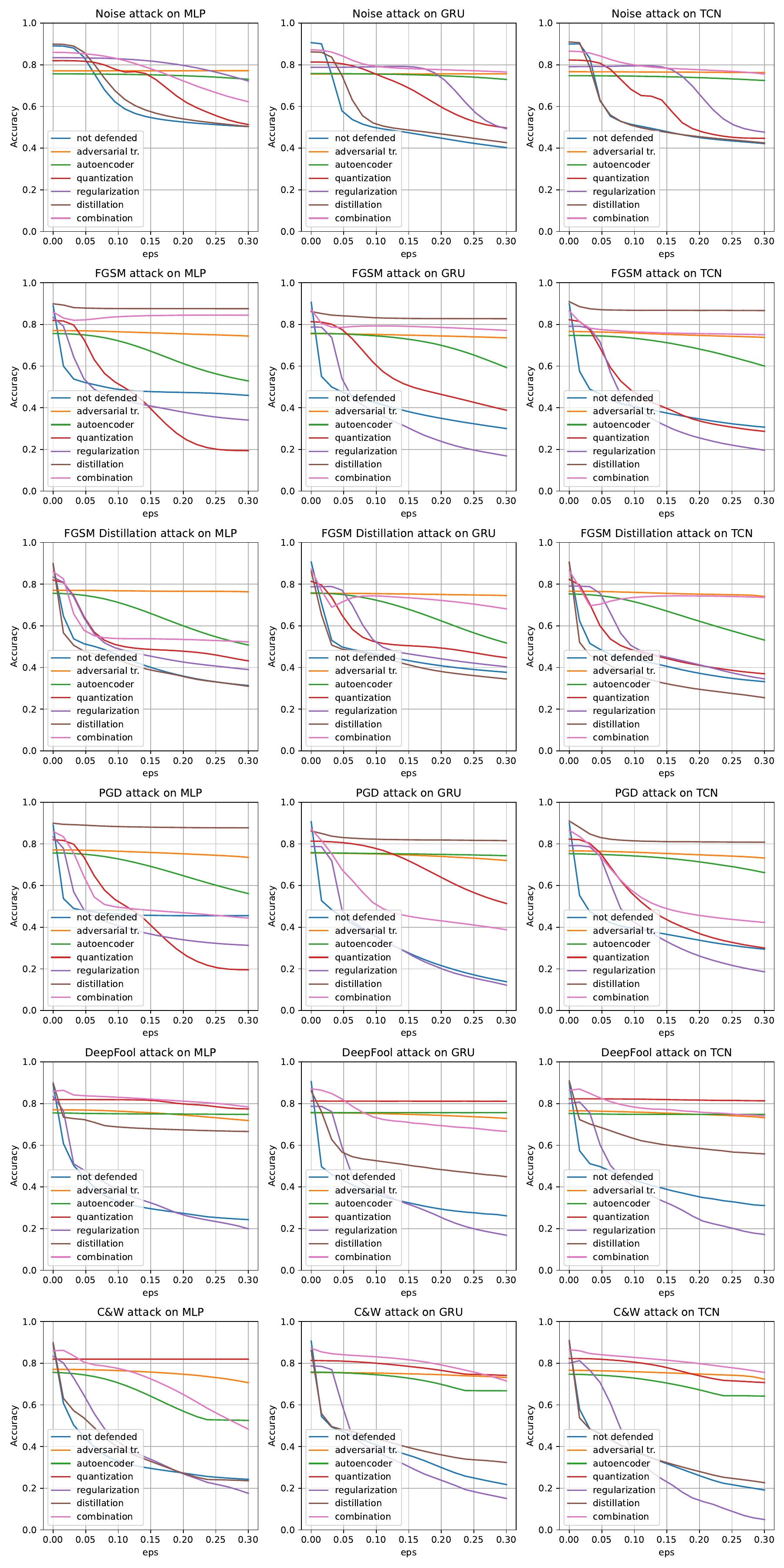}
    \caption{Accuracy of protected and unprotected models after adversarial attacks.}
    \label{fig:all}
\end{figure*}

In this section we present the final comparison of all combinations of models, attacks and defenses (Tables \ref{table:mlp}, \ref{table:gru}, \ref{table:tcn} and Fig. \ref{fig:all}).

\begin{table*}[!ht]
\tiny
\caption{Accuracy of protected and unprotected MLP model after adversarial attacks with different $\epsilon$ values.}
\label{table:mlp}
\centering
\begin{tabular}{ c|c|c|c|c|c|c|c|c|c|c|c|c } 
 \multicolumn{2}{c|}{$\epsilon$} &0.015&0.03&0.06&0.09&0.12&0.15&0.18&0.21&0.24&0.27&0.30\\
 \hline\hline
 Noise attack&Unprotected & \textbf{0.89} & \textbf{0.88} & 0.76 & 0.62 & 0.57 & 0.54 & 0.53 & 0.52 & 0.52 & 0.51 & 0.50 \\
 &Adversarial training & 0.77 & 0.77 & 0.77 & 0.77 & 0.77 & \textbf{0.77} & \textbf{0.77} & \textbf{0.77} & \textbf{0.77} & \textbf{0.77} & \textbf{0.77} \\
 &Autoencoder & 0.76 & 0.76 & 0.76 & 0.75 & 0.75 & 0.75 & 0.75 & 0.75 & 0.74 & 0.74 & \textbf{0.73} \\
 &Quantization & 0.82 & 0.82 & 0.81 & 0.78 & 0.77 & 0.73 & 0.69 & 0.62 & 0.57 & 0.54 & 0.51 \\
 &Regularization & 0.83 & 0.83 & \textbf{0.83} & \textbf{0.83} & \textbf{0.82} & \textbf{0.82} & \textbf{0.81} & \textbf{0.79} & \textbf{0.77} & \textbf{0.75} & 0.72 \\
 &Distillation & \textbf{0.90} & \textbf{0.89} & 0.80 & 0.67 & 0.60 & 0.57 & 0.56 & 0.54 & 0.52 & 0.51 & 0.50 \\
 &Combination (ours) & 0.86 & 0.86 & \textbf{0.85} & \textbf{0.83} & \textbf{0.81} & \textbf{0.77} & 0.75 & 0.72 & 0.68 & 0.65 & 0.62 \\
 \hline
 FGSM attack&Unprotected & 0.60 & 0.54 & 0.51 & 0.49 & 0.48 & 0.48 & 0.48 & 0.47 & 0.47 & 0.47 & 0.46 \\
 &Adversarial training & 0.77 & 0.77 & 0.77 & 0.77 & 0.76 & 0.76 & 0.76 & 0.75 & 0.75 & 0.75 & 0.74 \\
 &Autoencoder & 0.76 & 0.75 & 0.74 & 0.73 & 0.70 & 0.66 & 0.64 & 0.61 & 0.57 & 0.55 & 0.53 \\
 &Quantization & 0.82 & 0.80 & 0.63 & 0.52 & 0.46 & 0.37 & 0.32 & 0.25 & 0.21 & 0.20 & 0.19 \\
 &Regularization & 0.79 & 0.64 & 0.49 & 0.45 & 0.42 & 0.40 & 0.39 & 0.38 & 0.36 & 0.35 & 0.34 \\
 &Distillation & \textbf{0.89} & \textbf{0.88} & \textbf{0.88} & \textbf{0.88} & \textbf{0.88} & \textbf{0.88} & \textbf{0.88} & \textbf{0.88} & \textbf{0.88} & \textbf{0.88} & \textbf{0.88} \\
 &Combination (ours) & \textbf{0.83} & \textbf{0.82} & \textbf{0.83} & \textbf{0.84} & \textbf{0.84} & \textbf{0.84} & \textbf{0.84} & \textbf{0.84} & \textbf{0.84} & \textbf{0.84} & \textbf{0.84} \\
 \hline
 FGSM distillation attack&Unprotected & 0.65 & 0.54 & 0.50 & 0.47 & 0.43 & 0.39 & 0.38 & 0.36 & 0.34 & 0.32 & 0.31 \\
 &Adversarial training & 0.77 & \textbf{0.77} & \textbf{0.77} & \textbf{0.77} & \textbf{0.77} & \textbf{0.77} & \textbf{0.77} & \textbf{0.77} & \textbf{0.77} & \textbf{0.77} & \textbf{0.76} \\
 &Autoencoder & 0.75 & \textbf{0.75} & \textbf{0.74} & \textbf{0.72} & \textbf{0.69} & \textbf{0.65} & \textbf{0.63} & \textbf{0.59} & \textbf{0.56} & \textbf{0.53} & 0.51 \\
 &Quantization & \textbf{0.81} & 0.74 & 0.57 & 0.51 & 0.49 & 0.49 & 0.48 & 0.48 & 0.47 & 0.45 & 0.43 \\
 &Regularization & \textbf{0.81} & 0.74 & 0.56 & 0.49 & 0.47 & 0.45 & 0.44 & 0.42 & 0.41 & 0.40 & 0.39 \\
 &Distillation & 0.57 & 0.51 & 0.46 & 0.44 & 0.41 & 0.38 & 0.37 & 0.35 & 0.34 & 0.32 & 0.31 \\
 &Combination (ours) & \textbf{0.83} & 0.66 & 0.55 & 0.54 & 0.54 & 0.54 & 0.54 & 0.53 & 0.53 & \textbf{0.53} & \textbf{0.52} \\
\hline
 PGD attack&Unprotected & 0.54 & 0.49 & 0.47 & 0.46 & 0.46 & 0.46 & 0.46 & 0.46 & 0.46 & 0.46 & 0.46 \\
 &Adversarial training & 0.77 & 0.77 & \textbf{0.77} & \textbf{0.77} & \textbf{0.76} & \textbf{0.76} & \textbf{0.76} & \textbf{0.75} & \textbf{0.75} & \textbf{0.74} & \textbf{0.74} \\
 &Autoencoder & 0.76 & 0.75 & 0.75 & 0.73 & 0.71 & 0.68 & 0.67 & 0.64 & 0.61 & 0.59 & 0.56 \\
 &Quantization & 0.82 & \textbf{0.80} & 0.65 & 0.53 & 0.74 & 0.38 & 0.34 & 0.26 & 0.22 & 0.20 & 0.20 \\
 &Regularization & 0.78 & 0.57 & 0.45 & 0.41 & 0.38 & 0.36 & 0.35 & 0.34 & 0.33 & 0.32 & 0.31 \\
 &Distillation & \textbf{0.89} & \textbf{0.89} & \textbf{0.89} & \textbf{0.88} & \textbf{0.88} & \textbf{0.88} & \textbf{0.88} & \textbf{0.88} & \textbf{0.88} & \textbf{0.88} & \textbf{0.88} \\
 &Combination (ours) & \textbf{0.83} & 0.76 & 0.54 & 0.50 & 0.49 & 0.48 & 0.48 & 0.47 & 0.46 & 0.45 & 0.44 \\
 \hline
 DeepFool attack&Unprotected & 0.61 & 0.50 & 0.40 & 0.34 & 0.31 & 0.29 & 0.28 & 0.27 & 0.26 & 0.25 & 0.24 \\
 &Adversarial training & 0.77 & 0.77 & 0.77 & 0.76 & 0.76 & 0.75 & 0.75 & 0.74 & 0.74 & 0.73 & 0.72 \\
 &Autoencoder & 0.75 & 0.75 & 0.75 & 0.75 & 0.75 & 0.75 & 0.75 & 0.75 & 0.75 & 0.75 & 0.75 \\
 &Quantization & \textbf{0.82} & \textbf{0.82} & \textbf{0.82} & \textbf{0.82} & \textbf{0.82} & \textbf{0.81} & \textbf{0.81} & \textbf{0.8} & \textbf{0.79} & \textbf{0.78} & \textbf{0.77} \\
 &Regularization & 0.77 & 0.51 & 0.45 & 0.39 & 0.35 & 0.32 & 0.3 & 0.26 & 0.24 & 0.22 & 0.2 \\
 &Distillation & 0.73 & 0.73 & 0.71 & 0.69 & 0.68 & 0.68 & 0.68 & 0.67 & 0.67 & 0.67 & 0.67 \\
 &Combination (ours) & \textbf{0.86} & \textbf{0.84} & \textbf{0.84} & \textbf{0.83} & \textbf{0.83} & \textbf{0.82} & \textbf{0.82} & \textbf{0.81} & \textbf{0.80} & \textbf{0.80} & \textbf{0.78} \\
 \hline
 C\&W attack&Unprotected & 0.61 & 0.50 & 0.40 & 0.34 & 0.31 & 0.29 & 0.28 & 0.27 & 0.26 & 0.25 & 0.24 \\
 &Adversarial training & 0.77 & 0.77 & 0.77 & 0.77 & \textbf{0.76} & \textbf{0.76} & \textbf{0.75} & \textbf{0.75} & \textbf{0.74} & \textbf{0.72} & \textbf{0.71} \\
 &Autoencoder & 0.75 & 0.75 & 0.74 & 0.71 & 0.67 & 0.63 & 0.61 & 0.56 & 0.53 & 0.53 & 0.53 \\
 &Quantization & \textbf{0.82} & \textbf{0.82} & \textbf{0.82} & \textbf{0.82} & \textbf{0.82} & \textbf{0.82} & \textbf{0.82} & \textbf{0.82} & \textbf{0.82} & \textbf{0.82} & \textbf{0.82} \\
 &Regularization & 0.8 & 0.73 & 0.56 & 0.43 & 0.37 & 0.33 & 0.31 & 0.26 & 0.23 & 0.21 & 0.18 \\
 &Distillation & 0.63 & 0.57 & 0.49 & 0.41 & 0.36 & 0.32 & 0.3 & 0.27 & 0.24 & 0.24 & 0.24 \\
 &Combination (ours) & \textbf{0.86} & \textbf{0.84} & \textbf{0.79} & \textbf{0.78} & 0.75 & 0.71 & 0.69 & 0.64 & 0.58 & 0.53 & 0.48 \\
\end{tabular}
\end{table*}

\begin{table*}[!ht]
\tiny
\caption{Accuracy of protected and unprotected GRU model after adversarial attacks with different $\epsilon$ values.}
\label{table:gru}
\centering
\begin{tabular}{ c|c|c|c|c|c|c|c|c|c|c|c|c } 
 \multicolumn{2}{c|}{$\epsilon$} &0.015&0.03&0.06&0.09&0.12&0.15&0.18&0.21&0.24&0.27&0.30\\
 \hline\hline
 Noise attack&Unprotected & \textbf{0.90} & 0.75 & 0.54 & 0.50 & 0.48 & 0.47 & 0.46 & 0.44 & 0.43 & 0.41 & 0.40 \\
 &Adversarial training & 0.76 & 0.76 & 0.76 & 0.76 & 0.76 & 0.76 & 0.76 & \textbf{0.76} & \textbf{0.76} & \textbf{0.76} & \textbf{0.76} \\
 &Autoencoder & 0.76 & 0.76 & 0.76 & 0.76 & 0.75 & 0.75 & 0.75 & 0.75 & 0.74 & 0.74 & 0.73 \\
 &Quantization & 0.81 & 0.81 & \textbf{0.80} & 0.76 & 0.72 & 0.68 & 0.65 & 0.59 & 0.54 & 0.51 & 0.50 \\
 &Regularization & 0.79 & 0.79 & 0.79 & \textbf{0.79} & \textbf{0.79} & \textbf{0.79} & \textbf{0.78} & 0.73 & 0.62 & 0.53 & 0.49 \\
 &Distillation & 0.86 & \textbf{0.84} & 0.63 & 0.52 & 0.50 & 0.48 & 0.48 & 0.47 & 0.45 & 0.44 & 0.43 \\
 &Combination (ours) & \textbf{0.87} & \textbf{0.86} & \textbf{0.82} & \textbf{0.79} & \textbf{0.79} & \textbf{0.78} & \textbf{0.78} & \textbf{0.78} & \textbf{0.77} & \textbf{0.77} & \textbf{0.77} \\
 \hline
 FGSM attack&Unprotected & 0.55 & 0.50 & 0.46 & 0.43 & 0.40 & 0.38 & 0.36 & 0.35 & 0.33 & 0.31 & 0.30 \\
 &Adversarial training & 0.76 & 0.76 & 0.75 & 0.75 & 0.75 & 0.75 & 0.75 & 0.74 & 0.74 & 0.74 & 0.74 \\
 &Autoencoder & 0.76 & 0.76 & 0.75 & 0.75 & 0.74 & 0.73 & 0.72 & 0.70 & 0.67 & 0.63 & 0.59 \\
 &Quantization & \textbf{0.81} & \textbf{0.80} & 0.73 & 0.62 & 0.54 & 0.50 & 0.48 & 0.46 & 0.44 & 0.41 & 0.39 \\
 &Regularization & 0.79 & 0.74 & 0.45 & 0.38 & 0.34 & 0.29 & 0.27 & 0.23 & 0.21 & 0.19 & 0.17 \\
 &Distillation & \textbf{0.85} & \textbf{0.84} & \textbf{0.84} & \textbf{0.83} & \textbf{0.83} & \textbf{0.83} & \textbf{0.83} & \textbf{0.83} & \textbf{0.83} & \textbf{0.83} & \textbf{0.83} \\
 &Combination (ours) & \textbf{0.81} & 0.79 & \textbf{0.79} & \textbf{0.79} & \textbf{0.79} & \textbf{0.79} & \textbf{0.79} & \textbf{0.79} & \textbf{0.78} & \textbf{0.78} & \textbf{0.77} \\
 \hline
 FGSM distillation attack&Unprotected & 0.71 & 0.53 & 0.49 & 0.47 & 0.45 & 0.43 & 0.42 & 0.41 & 0.39 & 0.39 & 0.28 \\
 &Adversarial training & 0.76 & \textbf{0.76} & \textbf{0.76} & \textbf{0.75} & \textbf{0.75} & \textbf{0.75} & \textbf{0.75} & \textbf{0.75} & \textbf{0.75} & \textbf{0.75} & \textbf{0.75} \\
 &Autoencoder & 0.76 & 0.75 & \textbf{0.74} & 0.73 & 0.70 & 0.67 & 0.65 & 0.62 & 0.58 & 0.55 & 0.52 \\
 &Quantization & \textbf{0.80} & 0.74 & 0.58 & 0.53 & 0.51 & 0.50 & 0.50 & 0.49 & 0.48 & 0.46 & 0.45 \\
 &Regularization & \textbf{0.79} & \textbf{0.79} & 0.70 & 0.53 & 0.48 & 0.46 & 0.45 & 0.44 & 0.43 & 0.41 & 0.40 \\
 &Distillation & 0.65 & 0.51 & 0.48 & 0.46 & 0.43 & 0.41 & 0.40 & 0.38 & 0.37 & 0.35 & 0.34 \\
 &Combination (ours) & 0.77 & 0.69 & \textbf{0.74} & \textbf{0.74} & \textbf{0.74} & \textbf{0.73} & \textbf{0.73} & \textbf{0.72} & \textbf{0.71} & \textbf{0.70} & \textbf{0.68} \\
\hline
 PGD attack&Unprotected & 0.53 & 0.48 & 0.42 & 0.36 & 0.31 & 0.26 & 0.24 & 0.21 & 0.18 & 0.16 & 0.14 \\
 &Adversarial training & 0.76 & 0.76 & 0.75 & 0.75 & \textbf{0.75} & \textbf{0.75} & 0.74 & 0.74 & 0.73 & 0.73 & 0.72 \\
 &Autoencoder & 0.76 & 0.76 & 0.76 & 0.75 & \textbf{0.75} & \textbf{0.75} & \textbf{0.75} & \textbf{0.75} & \textbf{0.75} & \textbf{0.75} & \textbf{0.74} \\
 &Quantization & 0.81 & \textbf{0.81} & \textbf{0.80} & \textbf{0.78} & \textbf{0.75} & 0.71 & 0.68 & 0.63 & 0.58 & 0.54 & 0.51 \\
 &Regularization & 0.79 & 0.72 & 0.43 & 0.36 & 0.31 & 0.26 & 0.23 & 0.19 & 0.17 & 0.14 & 0.12 \\
 &Distillation & \textbf{0.85} & \textbf{0.84} & \textbf{0.83} & \textbf{0.82} & \textbf{0.82} & \textbf{0.82} & \textbf{0.82} & \textbf{0.82} & \textbf{0.82} & \textbf{0.82} & \textbf{0.82} \\
 &Combination (ours) & \textbf{0.82} & 0.75 & 0.62 & 0.52 & 0.47 & 0.45 & 0.44 & 0.43 & 0.42 & 0.40 & 0.39 \\
 \hline
 DeepFool attack&Unprotected & 0.50 & 0.46 & 0.41 & 0.37 & 0.34 & 0.32 & 0.31 & 0.29 & 0.28 & 0.27 & 0.26 \\
 &Adversarial training & 0.76 & 0.76 & 0.75 & 0.75 & 0.75 & 0.75 & 0.75 & 0.74 & 0.74 & 0.73 & 0.73 \\
 &Autoencoder & 0.76 & 0.76 & 0.76 & \textbf{0.76} & \textbf{0.76} & \textbf{0.76} & \textbf{0.76} & \textbf{0.76} & \textbf{0.76} & \textbf{0.76} & \textbf{0.76} \\
 &Quantization & \textbf{0.81} & \textbf{0.81} & \textbf{0.81} & \textbf{0.81} & \textbf{0.81} & \textbf{0.81} & \textbf{0.81} & \textbf{0.81} & \textbf{0.81} & \textbf{0.81} & \textbf{0.81} \\
 &Regularization & 0.79 & 0.76 & 0.45 & 0.39 & 0.34 & 0.31 & 0.29 & 0.25 & 0.21 & 0.19 & 0.17 \\
 &Distillation & 0.76 & 0.63 & 0.54 & 0.53 & 0.51 & 0.50 & 0.49 & 0.48 & 0.47 & 0.46 & 0.45 \\
 &Combination (ours) & \textbf{0.86} & \textbf{0.85} & \textbf{0.79} & 0.74 & 0.72 & 0.71 & 0.70 & 0.69 & 0.68 & 0.68 & 0.67 \\
 \hline
 C\&W attack&Unprotected & 0.54 & 0.49 & 0.45 & 0.41 & 0.38 & 0.35 & 0.33 & 0.29 & 0.26 & 0.24 & 0.22 \\
 &Adversarial training & 0.76 & 0.76 & 0.75 & 0.75 & 0.75 & 0.75 & 0.75 & 0.74 & 0.74 & \textbf{0.74} & \textbf{0.73} \\
 &Autoencoder & 0.76 & 0.76 & 0.75 & 0.75 & 0.74 & 0.72 & 0.72 & 0.69 & 0.67 & 0.67 & 0.67 \\
 &Quantization & \textbf{0.81} & \textbf{0.81} & \textbf{0.81} & \textbf{0.80} & \textbf{0.79} & \textbf{0.78} & \textbf{0.78} & \textbf{0.76} & \textbf{0.75} & \textbf{0.75} & \textbf{0.74} \\
 &Regularization & 0.78 & 0.77 & 0.45 & 0.39 & 0.34 & 0.29 & 0.27 & 0.23 & 0.19 & 0.17 & 0.15 \\
 &Distillation & 0.56 & 0.50 & 0.47 & 0.44 & 0.42 & 0.39 & 0.38 & 0.36 & 0.34 & 0.33 & 0.32 \\
 &Combination (ours) & \textbf{0.85} & \textbf{0.85} & \textbf{0.84} & \textbf{0.83} & \textbf{0.82} & \textbf{0.81} & \textbf{0.81} & \textbf{0.79} & \textbf{0.77} & \textbf{0.74} & 0.72 \\
\end{tabular}
\end{table*}

\begin{table*}[!ht]
\tiny
\caption{Accuracy of protected and unprotected TCN model after adversarial attacks with different $\epsilon$ values.}
\label{table:tcn}
\centering
\begin{tabular}{ c|c|c|c|c|c|c|c|c|c|c|c|c } 
 \multicolumn{2}{c|}{$\epsilon$} &0.015&0.03&0.06&0.09&0.12&0.15&0.18&0.21&0.24&0.27&0.30\\
 \hline\hline
 Noise attack&Unprotected & \textbf{0.89} & \textbf{0.83} & 0.55 & 0.51 & 0.49 & 0.47 & 0.46 & 0.45 & 0.44 & 0.43 & 0.42 \\
 &Adversarial training & 0.77 & 0.77 & 0.77 & 0.77 & 0.77 & 0.76 & \textbf{0.76} & \textbf{0.76} & \textbf{0.76} & \textbf{0.76} & \textbf{0.76} \\
 &Autoencoder & 0.75 & 0.75 & 0.75 & 0.75 & 0.74 & 0.74 & 0.74 & 0.74 & 0.73 & 0.73 & 0.72 \\
 &Quantization & 0.82 & 0.82 & 0.78 & 0.68 & 0.65 & 0.57 & 0.52 & 0.48 & 0.46 & 0.45 & 0.45 \\
 &Regularization & 0.79 & 0.79 & \textbf{0.79} & \textbf{0.79} & \textbf{0.79} & \textbf{0.78} & 0.74 & 0.64 & 0.54 & 0.49 & 0.48 \\
 &Distillation & \textbf{0.91} & 0.79 & 0.56 & 0.51 & 0.49 & 0.47 & 0.46 & 0.45 & 0.44 & 0.43 & 0.42 \\
 &Combination (ours) & 0.86 & \textbf{0.86} & \textbf{0.83} & \textbf{0.80} & \textbf{0.79} & \textbf{0.78} & \textbf{0.78} & \textbf{0.78} & \textbf{0.77} & \textbf{0.76} & \textbf{0.76} \\
 \hline
 FGSM attack&Unprotected & 0.57 & 0.49 & 0.44 & 0.41 & 0.39 & 0.37 & 0.36 & 0.34 & 0.33 & 0.32 & 0.31 \\
 &Adversarial training & 0.77 & 0.77 & 0.76 & 0.76 & \textbf{0.76} & 0.75 & 0.75 & 0.75 & 0.74 & 0.74 & 0.74 \\
 &Autoencoder & 0.75 & 0.75 & 0.74 & 0.74 & 0.72 & 0.71 & 0.7 & 0.68 & 0.65 & 0.63 & 0.6 \\
 &Quantization & \textbf{0.81} & 0.77 & 0.59 & 0.49 & 0.43 & 0.39 & 0.36 & 0.33 & 0.31 & 0.3 & 0.29 \\
 &Regularization & 0.79 & \textbf{0.78} & 0.57 & 0.41 & 0.35 & 0.3 & 0.28 & 0.25 & 0.23 & 0.21 & 0.2 \\
 &Distillation & \textbf{0.89} & \textbf{0.88} & \textbf{0.87} & \textbf{0.87} & \textbf{0.87} & \textbf{0.87} & \textbf{0.87} & \textbf{0.87} & \textbf{0.87} & \textbf{0.87} & \textbf{0.87} \\
 &Combination (ours) & \textbf{0.81} & \textbf{0.78} & \textbf{0.77} & \textbf{0.77} & \textbf{0.76} & \textbf{0.76} & \textbf{0.76} & \textbf{0.76} & \textbf{0.75} & \textbf{0.75} & \textbf{0.75} \\
 \hline
 FGSM distillation attack&Unprotected & 0.62 & 0.52 & 0.46 & 0.44 & 0.42 & 0.40 & 0.39 & 0.37 & 0.35 & 0.34 & 0.33 \\
 &Adversarial training & 0.77 & \textbf{0.77} & \textbf{0.76} & \textbf{0.76} & \textbf{0.76} & \textbf{0.76} & \textbf{0.75} & \textbf{0.75} & \textbf{0.75} & \textbf{0.75} & \textbf{0.74} \\
 &Autoencoder & 0.75 & 0.75 & \textbf{0.74} & 0.72 & 0.69 & 0.66 & 0.65 & 0.62 & 0.59 & 0.56 & 0.53 \\
 &Quantization & \textbf{0.8} & 0.7 & 0.54 & 0.49 & 0.46 & 0.44 & 0.43 & 0.41 & 0.39 & 0.38 & 0.37 \\
 &Regularization & \textbf{0.79} & \textbf{0.79} & 0.66 & 0.51 & 0.47 & 0.44 & 0.43 & 0.41 & 0.38 & 0.36 & 0.35 \\
 &Distillation & 0.52 & 0.46 & 0.41 & 0.37 & 0.34 & 0.32 & 0.31 & 0.29 & 0.28 & 0.27 & 0.26 \\
 &Combination (ours) & 0.78 & 0.70 & 0.72 & \textbf{0.73} & \textbf{0.74} & \textbf{0.74} & \textbf{0.74} & \textbf{0.74} & \textbf{0.74} & \textbf{0.74} & \textbf{0.74} \\
\hline
 PGD attack&Unprotected & 0.55 & 0.47 & 0.42 & 0.40 & 0.38 & 0.36 & 0.35 & 0.33 & 0.32 & 0.30 & 0.29 \\
 &Adversarial training & 0.77 & 0.77 & \textbf{0.76} & \textbf{0.76} & \textbf{0.76} & \textbf{0.75} & \textbf{0.75} & \textbf{0.75} & \textbf{0.74} & \textbf{0.74} & \textbf{0.73} \\
 &Autoencoder & 0.75 & 0.75 & 0.75 & 0.74 & 0.74 & 0.73 & 0.72 & 0.71 & 0.7 & 0.68 & 0.66 \\
 &Quantization & 0.82 & \textbf{0.8} & 0.69 & 0.57 & 0.48 & 0.42 & 0.4 & 0.36 & 0.34 & 0.32 & 0.3 \\
 &Regularization & 0.79 & 0.79 & 0.61 & 0.43 & 0.37 & 0.32 & 0.29 & 0.26 & 0.23 & 0.2 & 0.19 \\
 &Distillation & \textbf{0.88} & \textbf{0.85} & \textbf{0.82} & \textbf{0.82} & \textbf{0.81} & \textbf{0.81} & \textbf{0.81} & \textbf{0.81} & \textbf{0.81} & \textbf{0.81} & \textbf{0.81} \\
 &Combination (ours) & \textbf{0.84} & 0.79 & 0.68 & 0.58 & 0.52 & 0.48 & 0.47 & 0.45 & 0.44 & 0.43 & 0.42 \\
 \hline
 DeepFool attack&Unprotected & 0.57 & 0.51 & 0.48 & 0.44 & 0.41 & 0.38 & 0.37 & 0.35 & 0.33 & 0.32 & 0.31 \\
 &Adversarial training & 0.77 & 0.76 & 0.76 & 0.76 & 0.76 & 0.75 & 0.75 & 0.75 & 0.74 & 0.74 & 0.73 \\
 &Autoencoder & 0.75 & 0.75 & 0.75 & 0.75 & 0.75 & 0.75 & 0.75 & 0.75 & \textbf{0.75} & \textbf{0.75} & \textbf{0.75} \\
 &Quantization & \textbf{0.82} & \textbf{0.82} & \textbf{0.82} & \textbf{0.82} & \textbf{0.82} & \textbf{0.82} & \textbf{0.82} & \textbf{0.82} & \textbf{0.81} & \textbf{0.81} & \textbf{0.81} \\
 &Regularization & 0.81 & 0.75 & 0.5 & 0.41 & 0.36 & 0.31 & 0.29 & 0.24 & 0.21 & 0.19 & 0.17 \\
 &Distillation & 0.72 & 0.7 & 0.67 & 0.64 & 0.61 & 0.6 & 0.59 & 0.58 & 0.57 & 0.56 & 0.56 \\
 &Combination (ours) & \textbf{0.87} & \textbf{0.85} & \textbf{0.81} & \textbf{0.79} & \textbf{0.77} & \textbf{0.77} & \textbf{0.76} & \textbf{0.76} & \textbf{0.75} & \textbf{0.75} & 0.74 \\
 \hline
 C\&W attack&Unprotected & 0.58 & 0.48 & 0.43 & 0.39 & 0.35 & 0.31 & 0.29 & 0.25 & 0.22 & 0.21 & 0.19 \\
 &Adversarial training & 0.77 & 0.77 & 0.76 & 0.76 & 0.76 & 0.75 & 0.75 & \textbf{0.75} & \textbf{0.74} & \textbf{0.74} & \textbf{0.72} \\
 &Autoencoder & 0.75 & 0.74 & 0.74 & 0.73 & 0.72 & 0.7 & 0.69 & 0.67 & 0.64 & 0.64 & 0.64 \\
 &Quantization & \textbf{0.82} & \textbf{0.82} & \textbf{0.82} & \textbf{0.81} & \textbf{0.79} & \textbf{0.77} & \textbf{0.76} & 0.74 & 0.72 & 0.71 & 0.71 \\
 &Regularization & 0.81 & 0.77 & 0.61 & 0.4 & 0.28 & 0.22 & 0.18 & 0.14 & 0.1 & 0.07 & 0.05 \\
 &Distillation & 0.54 & 0.48 & 0.43 & 0.38 & 0.35 & 0.32 & 0.3 & 0.28 & 0.25 & 0.24 & 0.23 \\
 &Combination (ours) & \textbf{0.86} & \textbf{0.85} & \textbf{0.84} & \textbf{0.83} & \textbf{0.82} & \textbf{0.81} & \textbf{0.81} & \textbf{0.80} & \textbf{0.78} & \textbf{0.77} & \textbf{0.76} \\
\end{tabular}
\end{table*}

\bibliographystyle{IEEEtranIES}

\bibliography{reference}

\end{document}